\documentclass[11pt,a4paper,final]{article}

\usepackage[hyperref, acceptedWithA]{tacl2018v2}

\usepackage[lofdepth,lotdepth]{subfig}
\usepackage{graphicx}
\usepackage{multirow}
\usepackage{times,latexsym}
\usepackage{amsmath}
\usepackage{amsfonts}
\usepackage{enumerate}
\usepackage{amsthm}
\usepackage{url}
\usepackage{tikz}
\usepackage{tikz-dependency}
\usepackage{adjustbox}
\usepackage{booktabs}
\usepackage{mathtools}
\usepackage{xfrac}
\usepackage{xspace}
\usepackage{microtype}
\usepackage{bm}
\usepackage{emoji}

\newcommand\blfootnote[1]{%
  \begingroup
  \renewcommand\thefootnote{}\footnote{#1}%
  \addtocounter{footnote}{-1}%
  \endgroup
}

\makeatletter
\renewcommand*{\@fnsymbol}[1]{\ensuremath{\ifcase#1\or \emoji[twitter]{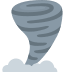}\or \dagger\or \ddagger\or
    \mathsection\or \mathparagraph\or \|\or **\or \dagger\dagger
    \or \ddagger\ddagger \else\@ctrerr\fi}}
\makeatother

\usepackage{color}

\usepackage{cleveref}
\crefname{section}{\S}{\S\S}
\Crefname{section}{\S}{\S\S}
\crefname{table}{Table}{}
\crefname{figure}{Figure}{}
\crefname{algorithm}{Algorithm}{}
\crefname{equation}{eq.}{}
\crefname{appendix}{Appendix}{}
\crefformat{section}{\S#2#1#3}  

\newcommand{\topic}[1]{}

\newcommand{\msc}{\textsc{msc}}
\newcommand{\fem}{\textsc{fem}}
\newcommand{\gen}{\textsc{gen}}
\newcommand{\nom}{\textsc{nom}}
\newcommand{\acc}{\textsc{acc}}
\newcommand{\dat}{\textsc{dat}}
\newcommand{\pl}{\textsc{pl}}
\newcommand{\sg}{\textsc{sg}}
\newcommand{\calN}{{\cal N}}

\newcommand{\calG}{{\cal G}}

\newcommand{\calC}{{\cal C}}
\newcommand{\calV}{{\cal V}}

\newcommand{\MI}{{\mathrm{MI}}}
\newcommand{\NMI}{{\mathrm{NMI}}}

\newcommand{\ent}{{\mathrm{H}}}

\newcommand{\word}[1]{\textit{#1}}
\newcommand{\XX}{six\xspace}

\newcommand{\Vadj}{V_\textsc{adj}}

\newcommand{\subj}{\textsc{nsubj}\xspace}
\newcommand{\dobj}{\textsc{dobj}\xspace}
\newcommand{\iobj}{\textsc{iobj}\xspace}

\newcommand{\comment}[1]{}

\usepackage{linguex}
\usepackage[safe]{tipa}

\title{On the Relationships Between the Grammatical Genders of Inanimate Nouns and Their Co-Occurring Adjectives and Verbs}

\author
  {
	\begin{tabular}{lllll}
	Adina Williams\raise1.0ex\hbox{\normalfont \normalsize \emoji[twitter]{1F32A}, \textschwa}\raise1.0ex\hbox{\normalfont \normalsize}
	& \textbf {Ryan Cotterell\raise1.0ex\hbox{\normalfont\normalsize \emoji[twitter]{1F32A}, \textipa{A,H}}\raise1.0ex\hbox{\normalfont\normalsize}} &
	Lawrence Wolf-Sonkin\raise1.0ex\hbox{\normalfont\normalsize \textipa{S}}
	\end{tabular} \\
	\begin{tabular}{lllll}
    \textbf{Dami{\'a}n Blasi}\raise1.0ex\hbox{\normalfont\normalsize \textipa{P}}
	& \textbf{Hanna Wallach}\raise1.0ex\hbox{\normalfont\normalsize \textipa{Z}}
	\end{tabular}\\
    \raise1.0ex\hbox{\normalfont\normalsize \textschwa} Facebook AI Research~\;~\raise1.0ex\hbox{\normalfont\normalsize \textipa{A}}ETH Z{\"u}rich ~\;~\raise1.0ex\hbox{\normalfont\normalsize \textipa{H}}University of Cambridge\\
    \raise1.0ex\hbox{\normalfont\normalsize \textipa{S}}Johns Hopkins University~\;~\raise1.0ex\hbox{\normalfont\normalsize \textipa{P}}Universit{\"a}t Z{\"u}rich~\;~\raise1.0ex\hbox{\normalfont\normalsize \textipa{Z}}Microsoft Research \\
	\texttt{adinawilliams@fb.com}~\;~\texttt{ryan.cotterell@inf.ethz.ch}~\;~\texttt{lawrencews@jhu.edu} \\
	\texttt{damian.blasi@uzh.ch}~\;~\texttt{wallach@microsoft.com}
}

\begin{document}
\maketitle
\begin{abstract}
We use large-scale corpora in six different gendered languages, along
with tools from NLP and information theory, to test whether there is a
relationship between the grammatical genders of inanimate nouns and
the adjectives used to describe those nouns. For all six languages, we
find that there is a statistically significant relationship. We also
find that there are statistically significant relationships between
the grammatical genders of inanimate nouns and the verbs that take
those nouns as direct objects, as indirect objects, and as
subjects. We defer a deeper investigation of these relationships for
future work.\looseness=-1
\end{abstract}

\section{Introduction}
In many languages, nouns possess grammatical genders. When a noun
refers to an animate object, its grammatical gender typically reflects
the biological sex or gender identity of that object
\citep{zubin1986,corbett1991gender,kramer2014}.\blfootnote{$^{\emoji[twitter]{1F32A}}$equal contribution in this scientific whirlwind} For example, in
German, the word for a boss is grammatically feminine when it refers
to a woman, but grammatically masculine when it refers to a
man---\word{Chef\textbf{in}} and \word{Chef}, respectively. But
inanimate nouns (i.e., nouns that refer to inanimate objects) also
possess grammatical genders. Any German speaker will tell you that the
word for a bridge, \word{Br{\"u}cke}, is grammatically feminine, even
though bridges have neither biological sexes nor gender
identities. Historically, the grammatical genders of inanimate nouns
have been considered more idiosyncratic and less meaningful than the
grammatical genders of animate nouns \citep{brugmann1889,
  bloomfield1933language, fox1990, aikhenvald2000}. However, some
cognitive scientists have reopened this discussion by using laboratory
experiments to test whether speakers of gendered languages reveal
gender stereotypes \citep{sera1994}---for example, and
most famously, when choosing adjectives to describe inanimate nouns
\citep{boroditsky2003sex}.\looseness=-1

Although laboratory experiments are highly informative, they typically
involve small sample sizes. In this paper, we therefore use
large-scale corpora and tools from NLP and information theory to test
whether there is a relationship between the grammatical genders of
inanimate nouns and the adjectives used to describe those
nouns. Specifically, we calculate the mutual information (MI)---a
measure of the mutual statistical dependence between two random
variables---between the grammatical genders of inanimate nouns and the
adjectives that describe them (i.e., share a dependency arc labeled
\textsc{amod}) using large-scale corpora in six different gendered
languages (specifically, German, Italian, Polish, Portuguese, Russian,
and Spanish). For all six languages, we find that the MI is
statistically significant, meaning that there is a
relationship.\looseness=-1

We also test whether there are relationships between the grammatical
genders of inanimate nouns and the verbs that take those nouns as
direct objects, as indirect objects, and as subjects. For all six
languages, we find that there are statistically significant
relationships for the verbs that take those nouns as direct objects
and as subjects. For five of the six languages, we also find that
there is statistically significant relationship for the verbs that
take those nouns as indirect objects, but because of the small number
of noun--verb pairs involved, we caution against reading too much into
this finding.

To contextualize our findings, we test whether there are statistically
significant relationships between the grammatical genders of inanimate
nouns and the cases and numbers of these nouns. A priori, we do not
expect to find statistically significant relationships, so these tests
can be viewed as a baseline of sorts. As expected, for each of the six
languages, there are no statistically significant relationships.

To provide further context, we also repeat all tests for
\emph{animate} nouns---a ``skyline'' of sorts---finding that for all
six languages there is a statistically significant relationship
between the grammatical genders of animate nouns and the adjectives
used to describe those nouns. We also find that there are
statistically significant relationships between the grammatical
genders of animate nouns and the verbs that take those nouns as direct
objects, as indirect objects, and as subjects. All of these
relationships have effect sizes (operationalized as normalized MI
values) that are larger than the effect sizes for inanimate
nouns.\looseness=-1

We emphasize that the practical significance and implications of our
findings require a deeper investigation. Most importantly, we do not
investigate the characteristics of the relationships that we
find. This means that we do not know whether these relationships are
characterized by gender stereotypes, as argued by some cognitive
scientists. We also do not engage with the ways that historical and
sociopolitical factors affect the grammatical genders possessed by
either animate or inanimate nouns
\citep{fodor1959,ibrahim2014}.\looseness=-1

\section{Background}\label{sec:background}
\subsection{Grammatical Gender}\label{sec:grammatical-gender}

Languages lie along a continuum with respect to whether nouns possess
grammatical genders. Languages with no grammatical genders, like
Turkish, lie on one end of this continuum, while languages with tens
of gender-like classes, like Swahili \cite{corbett1991gender}, lie on
the other. In this paper, we focus on \XX different gendered languages
for which large-scale corpora are readily available: German, Italian,
Polish, Portuguese, Russian, and Spanish---all languages of
Indo-European descent. Three of these languages (Italian, Portuguese,
and Spanish) have two grammatical genders---masculine and
feminine---while the other two have three grammatical
genders---masculine, feminine, and neuter.\looseness=-1

All six languages exhibit gender agreement, meaning that words are
marked with morphological suffixes that reflect the grammatical genders
of their surrounding nouns \cite{corbett2006agreement}. For example,
consider the following translations of the sentence, \emph{``The
  delicate fork is on the cold ground.''}\looseness=-1
\exg. \textit{\textbf{Die}} \textit{zierlich\textbf{e}} \textit{Gabel}
\textit{steht} \textit{auf} \textit{\textbf{dem}}
\textit{kalt\textbf{en}}
\textit{\textbf{Boden}}.\\ the.\textsc{f}.\textsc{sg}.\textsc{nom}
delicate.\textsc{f}.\textsc{sg}.\textsc{nom}
fork.\textsc{f}.\textsc{sg}.\textsc{nom} stands on
the.\textsc{m}.\textsc{sg}.\textsc{dat}
cold.\textsc{m}.\textsc{sg}.\textsc{dat}
ground.\textsc{m}.\textsc{sg}.\textsc{dat}\\ {The delicate fork is on
  the cold ground.}

\exg. \textit{\textbf{El}} \textit{tenedor} \textit{delicad\textbf{o}}
\textit{est{\'a}} \emph{en} \emph{\textbf{el}} \emph{suel\textbf{o}}
\textit{fr\'i\textbf{o}}.\\ the.\textsc{m}.\textsc{sg}
fork.\textsc{m}.\textsc{sg} delicate.\textsc{m}.\textsc{sg} is on
the.\textsc{m}.\textsc{sg} ground.\textsc{m}.\textsc{sg}
cold.\textsc{m}.\textsc{sg}\\ {The delicate fork is on the cold
  ground.}

Because the German word for a fork, \word{Gabel}, is grammatically
feminine, the German translation uses the feminine determiner,
\word{die}. Had \word{Gabel} been masculine, the German translation
would have used the masculine determiner, \word{der}. Similarly,
because the Spanish word for a fork, \word{tenedor}, is grammatically
masculine, the Spanish translation uses the masculine determiner,
\word{el}, instead of the feminine determiner, \word{la}. As we
explain in Section \ref{sec:preparation}, we lemmatize each corpus to
ensure that our tests do not simply reflect the presence of gender
agreement.

\subsection{Grammatical Gender \& Meaning}

Although some scholars have described the grammatical genders
possessed by inanimate nouns as ``creative'' and meaningful
\citep{grimm1890, wheeler1899}, many scholars have considered them to
be idiosyncratic \citep{brugmann1889,bloomfield1933language} or
arbitrary \citep[317]{maratsos1979}. In an overview of this work,
\newcite{dye2017functional} wrote, ``As often as not, the languages of
the world assign [inanimate] objects into seemingly arbitrary
[classes]\textellipsis \citeauthor{ockham1980} considered gender to be
a meaningless, unnecessary aspect of language.''
\newcite{bloomfield1933language} shared this viewpoint, stating that
``[t]here seems to be no practical criterion by which the gender of a
noun in German, French, or Latin [can] be determined.''  Indeed, adult
language learners often have particular difficulty mastering the
grammatical genders of inanimate nouns
(\citealt[Ch.4]{franceschina2005},
\citealt{dekeyser2005,montrul2008}), which suggests that their
meanings are not straightforward.\looseness=-1

Even if the grammatical genders possessed by inanimate nouns are
meaningless, ample evidence suggests that gender-related information
may affect cognitive processes \citep{sera1994,cubelli2005,
  cubelli2011, kurinski2011, boutonnet2012, saalbach2012}. Typologists
and formal linguists have argued that grammatical genders are an
important feature for morphosyntactic processes
\citep{corbett1991gender,corbett2006agreement,harbour2008,harbour2011,kramer2014,kramer2015},
while some cognitive scientists have shown that grammatical genders
can be a perceptual cue---for example, human brain responses exhibit
sensitivity to gender mismatches in several different languages
\citep{osterhout1995,hagoort1999,vigliocco2002,
  wicha2003,wicha2004,barber2004,barber2005,banon2012,caffarra2015},
and the grammatical genders of determiners and adjectives can prime
nouns \citep{bates1996,akhutina1999,friederici1999}. However, the
precise nature of the relationship between grammatical gender and
meaning remains an open research question.\looseness=-1

In particular, the grammatical genders possessed by inanimate nouns
might affect the ways that speakers of gendered languages
conceptualize the objects referred to by those nouns
\citep{jakobson1959,clarke1981,ervin1962,konishi1993,sera1994,sera2002,
  vigliocco2005grammatical,bassetti2007}---although we note that this
viewpoint is somewhat contentious
\citep{hofstatter1963,bender2011,mcwhorter2014language}. Neo-Whorfian cognitive scientists
hold a particularly strong variant of this viewpoint, arguing that
that the grammatical genders possessed by inanimate nouns prompt
speakers of gendered languages to rely on gender stereotypes when
choosing adjectives to describe those nouns \citep{Boroditsky00sex,
  boroditsky2002can, phillips2003can, boroditsky2003,
  boroditsky2003sex, semenuks2017}.  Most famously,
\newcite{boroditsky2003sex} claim to have conducted a laboratory
experiment showing that speakers of German choose stereotypically
feminine adjectives to describe, for example, bridges, while speakers
of Spanish choose stereotypically masculine adjectives, reflecting the
fact that in German, the word for a bridge, \word{Br{\"u}cke}, is
grammatically feminine, while in Spanish, the word for a bridge,
\word{puente}, is grammatically masculine. \newcite{boroditsky2003sex}
took these findings to be a relatively strong confirmation of the
existence of a stereotype effect---i.e., that speakers of gendered
languages reveal gender stereotypes when choosing adjectives to
describe inanimate nouns. That said, the experiment has not gone
unchallenged. Indeed, \newcite{mickan2014key} reported two
unsuccessful replication attempts. \looseness=-1

\subsection{Laboratory Experiments vs. Corpora}\label{sec:neo-whorphianism-gender-embeddings}

Traditionally, studies of grammatical gender and meaning have relied
on laboratory experiments. This is for two reasons: 1) laboratory
experiments can be tightly controlled, and 2) they enable scholars to
measure speakers' immediate, real-time speech production. However,
they also typically involve small sample sizes and, in many cases,
somewhat artificial settings. In contrast, large-scale corpora of
written text enable scholars to measure even relatively weak
correlations via writers' text production in natural, albeit less
tightly controlled, settings. They also facilitate the discovery of
correlations that hold across languages with disparate histories,
cultural contexts, and even gender systems. As a result, large-scale
corpora have proven useful for studying a wide variety of
language-related phenomena
\citep[e.g.,][]{featherston2007,kennedy2014,
  blasi2019distribution}.\looseness=-1

\begin{figure*}
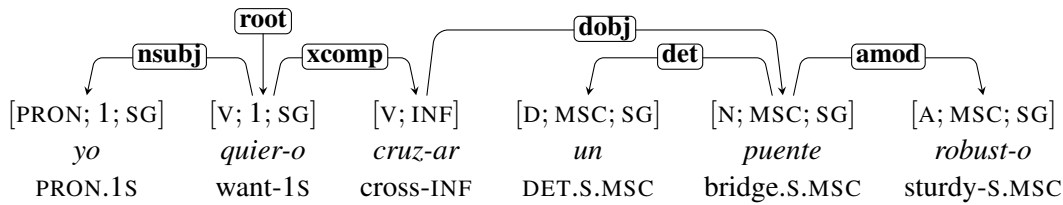

\centering
\begin{dependency}
\begin{deptext}[column sep=1em, row sep=.1ex]
$[\textsc{pron};\textsc{1};\textsc{sg}]$ \&
$[\textsc{v};\textsc{1};\textsc{sg}]$ \&
$[\textsc{v};\textsc{inf}]$ \&
$[\textsc{d};\textsc{msc};\textsc{sg}]$ \&
$[\textsc{n};\textsc{msc};\textsc{sg}]$ \&
$[\textsc{a};\textsc{msc};\textsc{sg}]$ \\
\it yo \& \it quier-o \& \it cruz-ar \& \it un \& \it puente \& \it robust-o \\
\textsc{pron}.1\textsc{s} \& want-1\textsc{s} \& cross-\textsc{inf} \& \textsc{det}.\textsc{s}.\textsc{msc} \& bridge.\textsc{s}.\textsc{msc} \& sturdy-\textsc{s}.\textsc{msc} \\
\end{deptext}
\depedge[style={font=\Large}]{2}{1}{\textbf{nsubj}}
\deproot[edge unit distance=1.75ex, style={font=\Large}]{2}{\textbf{root}}
\depedge[style={font=\Large}]{2}{3}{\textbf{xcomp}}
\depedge[style={font=\Large}]{5}{4}{\textbf{det}}
\depedge[edge unit distance=2.5ex, style={font=\Large}]{3}{5}{\textbf{dobj}}
\depedge[style={font=\Large}]{5}{6}{\textbf{amod}}
\end{dependency}
\caption{Dependency tree for the sentence, \word{"Yo quiero cruzar un puente robusto."}}
\label{fig:tree}
\end{figure*}

In this paper, we assume that a writer's choice of words in written
text is as informative as a speaker's choice of words in a laboratory
experiment, despite the obvious differences between these
settings. Consequently, we use large-scale corpora and tools from NLP
and information theory, enabling us to test for the presence of even
relatively weak relationships involving the grammatical genders of
inanimate nouns across multiple different gendered languages. We
therefore argue that our findings complement, rather than supersede,
laboratory experiments.\looseness=-1

\subsection{Related Work}

Our paper is not the first to use large-scale corpora and tools from
NLP to investigate gender and language. Many scholars have studied the
ways that societal norms and stereotypes, including gender norms and
stereotypes, can be reflected in representations of distributional
semantics derived from large-scale corpora, such as word embeddings
\citep{bolukbasi2016man,caliskan2017semantics, garg2018word,
  zhao-etal-2018-learning}. More recently, \newcite{williams2019}
found that the grammatical genders of inanimate nouns in eighteen
different languages were correlated with their lexical
semantics. \citet{dye2017functional} used tools from information
theory to reject the idea that the grammatical genders of nouns
separate those nouns into coherent categories, arguing instead that
grammatical genders are only meaningful in that they systematically
facilitate communication efficiency by reducing nominal entropy. Also
relevant to our paper is the work of \newcite{kann2019}, who proposed
a computational approach to testing whether there is a relationship
between the grammatical genders of inanimate nouns and the words that
co-occur with those nouns, operationalized via word
embeddings. However, in contrast to our findings, they found no
evidence for the presence of such a relationship. Finally, many
scholars have proposed a variety of computational techniques for
mitigating gender norms and stereotypes in a wide range of
language-based applications
\citep{dev2019attenuating,dinan2019queens,ethayarajh-etal-2019-understanding,hall-maudslay-etal-2019-name,
  stanovsky-etal-2019-evaluating, tan2019assessing,
  zhou-etal-2019-examining,zmigrod-etal-2019-counterfactual}.\looseness=-1

\section{Data Preparation}\label{sec:preparation}

We use the May, 2018 dump of Wikipedia to create a corpus for each of
the \XX different gendered languages (i.e., German, Italian, Polish,
Portuguese, Russian, and Spanish). Although Wikipedia is not the most
representative data source, this choice yields language-specific
corpora that are roughly parallel---i.e., they refer to the same
objects, but are not direct translations of each other (which could
lead to artificial word choices). We use UDPipe 1.0 to tokenize each
corpus
\citep{straka2016}.\footnote{\url{http://ufal.mff.cuni.cz/udpipe}}\looseness=-1

We dependency parse the corpus for each language using a
language-specific dependency parser \citep{andor2016, alberti2017},
trained using Universal Dependencies treebanks
\citep{11234/1-1983}. An example dependency tree is shown in
\cref{fig:tree}. We then extract all noun--adjective pairs (dependency
arcs labeled \textsc{amod}) and noun--verb pairs from each of the six
corpora; for verbs, we extract three types of pairs, reflecting the
fact that nouns can be direct objects (dependency arcs labeled \dobj),
indirect objects (dependency arcs labeled \iobj), or subjects
(dependency arcs labeled \subj) of verbs. We discard all pairs that
contain a noun that isn't present in
WordNet~\cite{princeton2010}.\footnote{\url{https://wordnet.princeton.edu/}}
We label the remaining nouns as ``animate'' or ``inanimate'' according
to WordNet.

Next, we lemmatize all words (i.e., nouns, adjectives, and
verbs). Each word is factored into a set of lexical features
consisting of a lemma, or canonical morphological form, and a bundle
of three morphological features corresponding to the grammatical
gender, number, and case of that word. For example, the German word
for a fork, \word{Gabel}, is grammatically feminine, singular, and
genitive. For nouns, we discard the lemmas themselves and retain only
the morphological features; for adjectives and verbs, we retain the
lemmas and discard the morphological features.\looseness=-1

For adjectives and verbs, lemmatizing is especially important because
it ensures that our tests do not simply reflect the presence of
gender agreement, as we describe in Section
\ref{sec:grammatical-gender}. However, this means that if the
lemmatizer fails, then our tests \emph{may} simply reflect
gender agreement despite our best efforts. To guard against this, we
use a state-of-the-art lemmatizer \cite{muller2015}, trained for each
language using Universal Dependencies treebanks
\citep{11234/1-1983}. We expect that when the lemmatizer fails, the
resulting lemmata will be low-frequency. We try to exclude
lemmatization failures from our calculations by discarding
low-frequency lemmata. For each language, we rank the adjective
lemmata by their token counts and retain only the highest-ranked
lemmata (in rank order) that account for 90\% of the adjective tokens;
we then discard all noun--adjective pairs that do not contain one of
these lemmata. We repeat the same process for verbs.\looseness=-1

Finally, to ensure that our tests reflect the most salient
relationships, we also discard low-frequency inanimate nouns and,
separately, low-frequency animate nouns using the same process. We
provide counts of the remaining noun--adjective and noun--verb pairs
in \cref{tab:inanimate_counts} (for inanimate nouns) and
\cref{tab:animate_counts} (for animate nouns).\looseness=-1

\section{Methodology}\label{sec:method}

For each language $\ell \in \{\textit{de},\textit{it}, \textit{pl},
\textit{pt}, \textit{ru}, \textit{es}\}$, we define
$\calV^{\ell}_{\textsc{adj}}$ to be the set of adjective lemmata
represented in the noun--adjective pairs retained for that language as
defined above. We similarly define $\calV^\ell_\textsc{verb}$ to be
the set of verb lemmata represented in the noun--verb pairs retained
for that language as described above. We then define
$\calV^\ell_{\textsc{verb-dobj}} \subset \calV^\ell_\textsc{verb}$,
$\calV^\ell_{\textsc{verb-iobj}} \subset \calV^\ell_\textsc{verb}$,
and $\calV^\ell_{\textsc{verb-subj}} \subset \calV^\ell_\textsc{verb}$
to be the sets of verbs that take the nouns as direct objects, as
indirect objects, and as subjects, respectively. We also define
$\calG^\ell$ to be the set of grammatical genders for that language
(e.g., $\calG^\textit{es} = \{\msc, \fem\}$), $\calC^\ell$ to be the
set of cases (e.g., $\calC^\textit{de} = \{\nom, \acc, \gen, \dat\}$),
and $\calN^\ell$ to be the set of numbers (e.g., $\calN^\textit{pt} =
\{\pl, \sg\}$). Finally, we define fourteen random variables:
$A^\ell_{i}$ and $A^\ell_{a}$ are $\calV^\ell_{\textsc{adj}}$-valued
random variables, $D_{i}^\ell$ and $D_a^\ell$ are
$\calV^\ell_{\textsc{verb-dobj}}$-valued random variables, $I^\ell_i$
and $I^\ell_a$ are $\calV^\ell_{\textsc{verb-iobj}}$-valued random
variables, $S^\ell_i$ and $S^\ell_a$ are
$\calV^\ell_{\textsc{verb-subj}}$-valued random variables, $G_i^\ell$
and $G_a^\ell$ are $\calG^\ell$-valued random variables, $C_i^\ell$
and $C_a^\ell$ are $\calC^\ell$-valued random variables, and
$N_i^\ell$ and $N_a^\ell$ are $\calN^\ell$-valued random
variables. The subscripts ``$i$'' and ``$a$'' denote inanimate and
animate nouns, respectively.\looseness=-1

To test whether there is a relationship between the grammatical
genders of inanimate nouns and the adjectives used to describe those
nouns for language $\ell$, we calculate the mutual information
(MI)---a measure of the mutual statistical dependence between two
random variables---between $G_i^\ell$ and $A_i^\ell$:
\begin{align}
  &\MI(G_i^\ell; A_i^\ell) \notag\\
  &\quad= \sum_{g \in \calG^\ell} \sum_{a \in \Vadj^\ell} P(g, a) \log_2 \frac{P_i(g, a)}{P_i(g)\, P_i(a)},
  \end{align}
where all probabilities are calculated with respect to inanimate nouns
only.  If $G_i^\ell$ and $A_i^\ell$ are independent---i.e., there is
no relationship between them---then $\MI(G_i^\ell; A_i^\ell) = 0$; if
$G_i^\ell$ and $A_i^\ell$ are maximally dependent then $\MI(G_i^\ell;
A_i^\ell) = {\min\{\ent(G_i^\ell), \ent(A_i^\ell)\}}$, where
$\ent(G_i^\ell)$ is the entropy of $G_i^\ell$ and $\ent(A_i^\ell)$ is
the entropy of $A_i^\ell$. For simplicity, we use plug-in estimates
for all probabilities (i.e., empirical probabilities), deferring the
use of more sophisticated estimators for future work. We note that
$\MI(G_i^\ell, A_i^\ell)$ can be calculated in ${\cal
  O}\left(|\calG^\ell|\cdot |\Vadj^\ell|\right)$ time; however,
$|\calG^\ell|$ is negligible (i.e, two or three) so the main cost is
$|\Vadj^\ell|$.

To test for statistical significance, we perform a permutation
test. Specifically, we permute the grammatical genders of the
inanimate nouns 10,000 times and, for each permutation, recalculate
the MI between $G_i^\ell$ and $A_i^\ell$ using the permuted
genders. We obtain a $p$-value by calculating the percentage of
permutations that have a higher MI than the MI obtained using the
non-permuted genders; if the $p$-value is less than $0.05$, then we
treat the relationship between $G_i^\ell$ and $A_i^\ell$ as
statistically significant.

Because the maximum possible MI between any pair of random variables
depends on the entropies of those variables, MI values are not
comparable across pairs of random variables. We therefore also
calculate the normalized MI (NMI) between $G_i^\ell$ and $A_i^\ell$ by
normalizing $\MI(G_i^\ell, A_i^\ell)$ to lie between zero and one. The
most obvious choice of normalizer is the maximum possible MI---i.e.,
${\min\{\ent(G_i^\ell), \ent(A_i^\ell)\}}$---however, various other
normalizers have been proposed, each of which has different advantages
and disadvantages \cite{gates2019}. We therefore calculate six
different variants of $\NMI(G_i^\ell, A_i^\ell)$ using the following
normalizers:\looseness=-1
\begin{gather}
\label{eq:norm1}
 {\min\{\ent(G_i^\ell), \ent(A_i^\ell)\}}\\
 \label{eq:norm2}
 \sqrt{\ent(G_i^\ell)\ent(A_i^\ell)}\\
 \label{eq:norm3}
 \frac{\ent(G_i^\ell) + \ent(A_i^\ell)}{2}\\
 \label{eq:norm4}
 {\max\{\ent(G_i^\ell), \ent(A_i^\ell)\}}\\
 \label{eq:norm5}
 {\max{\{\log{|\calG^\ell|}, \log{|\calV^\ell_{\textsc{adj}}|}\}}}\\
 \label{eq:norm6}
 \log{M_i^\ell},
 \end{gather}
where $M_i^\ell$ is the number of non-unique (inanimate)
noun--adjective pairs retained for that language.

To test whether there are relationships between the grammatical
genders of inanimate nouns and the verbs that take those nouns as
direct objects, as indirect objects, and as subjects, we calculate
$\MI(G_i^\ell, D_i^\ell)$, $\MI(G_i^\ell, I_i^\ell)$, and $\MI(G^\ell,
S_i^\ell)$. Again, all probabilities are calculated with respect to
inanimate nouns only, and we perform permutation tests to test for
statistical significance. We also calculate six NMI variants for each
of the three pairs of random variables, using normalizers that are
analogous to those in \Cref{eq:norm1} through \Cref{eq:norm6}.

As a baseline, we test whether there are relationships between the
grammatical genders of inanimate nouns and the cases and numbers of
those nouns---i.e., we calculate $\MI(G_i^\ell, C_i^\ell)$ and
$\MI(G_i^\ell, N^\ell_i)$ using probabilities that are calculated with
respect to inanimate nouns only. Again, we perform permutation tests
(but we do not expect that there will be statistically significant
relationships), and we calculate six NMI variants for each pair of
random variables using normalizers that are~analogous to those in
\Cref{eq:norm1} through \Cref{eq:norm6}.

Finally, we calculate $\MI(G_a^\ell, A_a^\ell)$, $\MI(G_a^\ell,
D_a^\ell)$, $\MI(G_a^\ell, I_a^\ell)$, $\MI(G_a^\ell, S_a^\ell)$,
$\MI(G_a^\ell, C_a^\ell)$, and $\MI(G_a^\ell, N_a^\ell)$) using
probabilities calculated with respect to \emph{animate} nouns
only. The first five of these are intended to serve as a ``skyline,''
while the last two are intended to serve as a sanity check (i.e., we
expect them to be close to zero, as with inanimate nouns). Again, we
perform permutation tests to test for statistical significance, and we
calculate six~NMI variants for each pair of random
variables.\looseness=-1

\section{Results}\label{sec:experiment1}

\begin{table*}[th]
  \centering
  \small
  \begin{tabular}{l  c c c c c c }
  \toprule
  & \textit{de} &
  \textit{it} &
  \textit{pl} &
  \textit{pt} &
  \textit{ru} &
  \textit{es}\\
  \midrule
  $\MI(G_i^\ell, A_i^\ell)$ & \textbf{0.0310} & \textbf{0.0500} & \textbf{0.0225} & \textbf{0.0400} & \textbf{0.0520} & \textbf{0.0664}\\
  $\MI(G_i^\ell, D_i^\ell)$ & \textbf{0.0290} & \textbf{0.0232} & \textbf{0.0109} & \textbf{0.0129} & \textbf{0.0440} & \textbf{0.0090}\\
  $\MI(G_i^\ell, I_i^\ell)$ & \textbf{0.0743} & 0.6973 & \textbf{0.0514} & \textbf{0.0230} & \textbf{0.0640} & \textbf{0.0184}\\
  $\MI(G_i^\ell, S_i^\ell)$ & \textbf{0.0276} & \textbf{0.0274} & \textbf{0.0226} & \textbf{0.0090} & \textbf{0.0270} & \textbf{0.0090}\\
      $\MI(G_i^\ell, C_i^\ell)$ & $<$ 0.001 & N/A & $<$ 0.001  & N/A & $<$ 0.001  &  N/A \\
      $\MI(G_i^\ell, N_i^\ell)$ & $<$ 0.001  &$<$ 0.001  & $<$0.001  & $<$0.001  & $<$ 0.001 & $<$ 0.001 \\
\bottomrule
  \end{tabular}
    \caption{The mutual information (MI) between the grammatical
      genders of inanimate nouns and a) the adjectives used to
      describe those nouns (top row), b) the verbs that take those
      nouns as direct objects, as indirect objects, and as subjects
      (rows 2--4, respectively), and c) the cases and numbers of those
      nouns (rows 5 and 6, respectively) for \XX different gendered
      languages. Statistical significance (i.e., a $p$-value less than
      0.05) is indicated using bold. MI values are not comparable
      across pairs of random variables.\looseness=-1}
    \vspace{1.0em}
  \label{tab:mi_significance}
  \end{table*}

In the first row of \cref{tab:mi_significance}, we provide the MI
between $G_i^\ell$ and $A_i^\ell$ for each language $\ell \in
\{\textit{de}, \textit{it}, \textit{pl}, \textit{pt}, \textit{ru},
\textit{es}\}$. For all six languages, $\MI (G_i^\ell, A_i^\ell)$ is
statistically significant (i.e., $p < 0.05$), meaning that there is a
relationship between the grammatical genders of inanimate nouns and
the adjectives used to describe those nouns. Rows 2--4 of
\cref{tab:mi_significance} contain $\MI(G_i^\ell,D_i^\ell)$,
$\MI(G_i^\ell,I_i^\ell)$, and $\MI(G^\ell,S_i^\ell)$ for each
language. For all six languages, $\MI(G_i^\ell,D_i^\ell)$ and
$\MI(G_i^\ell,S_i^\ell)$ are statistically significant (i.e., $p <
0.05$). For five of the six languages, $\MI(G_i^\ell,I_i^\ell)$ is
statistically significant, but because of the small number of
noun--verb pairs involved, we caution against reading too much into
this finding. We note that direct objects are closest to verbs in
analyses of constituent structures, followed by subjects and then
indirect objects \cite{chomsky1957, adger2003}. Finally, the last two
rows of \cref{tab:mi_significance} contain $\MI(G_i^\ell, C_i^\ell)$
and $\MI(G_i^\ell, N_i^\ell)$, respectively, for each language. We do
not find any statistically significant relationships for either case
or number.\looseness=-1

\begin{figure*}
  \centering
  \small
  \includegraphics[width=0.7\textwidth]{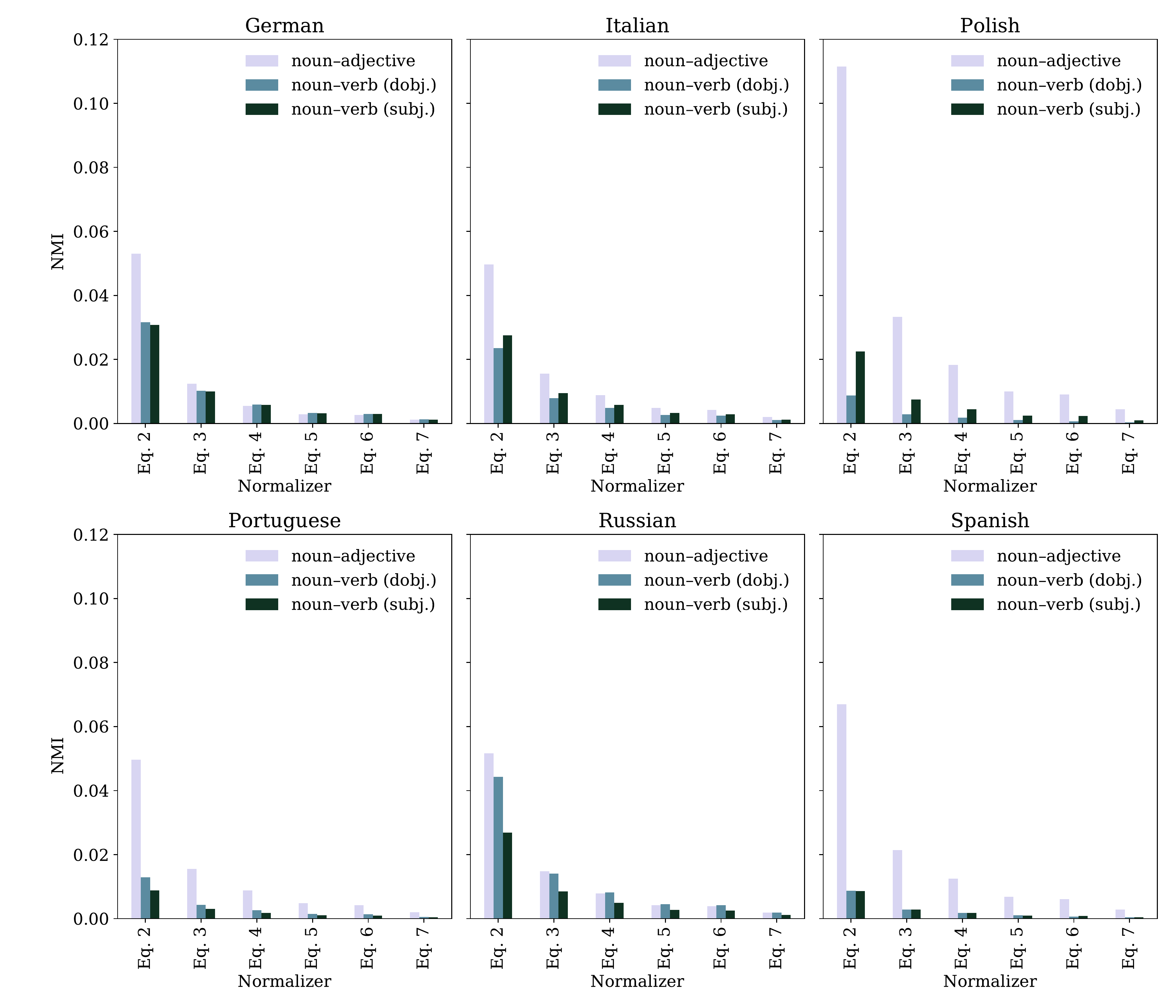}
  \caption{The normalized mutual information (NMI) between the
    grammatical genders of inanimate nouns and a) the adjectives used
    to describe those nouns b) the verbs that take those nouns as
    direct objects and as subjects for \XX different gendered
    languages. Each subplot contains six variants of~$\NMI(G_i^\ell,
    A_i^\ell)$, $\NMI(G_i^\ell,D_i^\ell)$, and
    $\NMI(G_i^\ell,S_i^\ell)$---one per normalizer---for a single
    language $\ell \in \{\textit{de}, \textit{it}, \textit{pl},
    \textit{pt}, \textit{ru}, \textit{es}\}$.}
  \label{fig:nmi_inanimates}
\end{figure*}

To facilitate comparisons, each subplot in \cref{fig:nmi_inanimates}
contains six variants of $\NMI(G_i^\ell, A_i^\ell)$,
$\NMI(G_i^\ell,D_i^\ell)$, and $\NMI(G_i^\ell,S_i^\ell)$, calculated
using normalizers that are analogous to those in \Cref{eq:norm1}
through \Cref{eq:norm6}, for a single language $\ell \in \{\textit{de},
\textit{it}, \textit{pl}, \textit{pt}, \textit{ru},
\textit{es}\}$. (We omit $\NMI(G_i^\ell,I_i^\ell)$ from each plot
because of the small number of noun--verb pairs involved.)
For $\ell \in \{ \textit{it},
\textit{pl}, \textit{pt}, \textit{es}\}$, $\NMI(G_i^\ell, A_i^\ell)$
is larger than $\NMI(G_i^\ell,D_i^\ell)$ and
$\NMI(G_i^\ell,S_i^\ell)$, regardless of the normalizer. For $\ell \in
\{\textit{it}, \textit{pl}\}$, $\NMI(G_i^\ell,S_i^\ell)$ is larger
than $\NMI(G_i^\ell,D_i^\ell)$;
$\NMI(G_i^\textit{pt},D_i^\textit{pt})$ is larger than
$\NMI(G_i^\textit{pt},S_i^\textit{pt})$; and
$\NMI(G_i^\textit{es},D_i^\textit{es})$ and
$\NMI(G_i^\textit{es},S_i^\textit{es})$ are roughly
comparable---again, all regardless of the normalizer. Meanwhile,
$\NMI(G_i^\textit{de}, A_i^\textit{de})$ is larger than
$\NMI(G_i^\textit{de},D_i^\textit{de})$ and
$\NMI(G_i^\textit{de},S_i^\textit{de})$ for the normalizer in
\Cref{eq:norm1}, while $\NMI(G_i^\textit{de}, A_i^\textit{de})$,
$\NMI(G_i^\textit{de},D_i^\textit{de})$, and
$\NMI(G_i^\textit{de},S_i^\textit{de})$ are all roughly comparable for
the other five normalizers. Finally, $\NMI(G_i^\textit{ru},
A_i^\textit{ru})$ and $\NMI(G_i^\textit{ru},D_i^\textit{ru})$ are
roughly comparable and larger than
$\NMI(G_i^\textit{ru},S_i^\textit{ru})$, regardless of the
normalizer.\looseness=-1

In other words, the relationship between the grammatical genders of
inanimate nouns and the adjectives used to describe those nouns is
generally stronger than, but sometimes roughly comparable to, the
relationships between the grammatical genders of inanimate nouns and
the verbs that take those nouns as direct objects and as
subjects. However, the relative strengths of the relationships between
the grammatical genders of inanimate nouns and the verbs that take
those nouns as direct objects and as subjects vary depending on the
language.\looseness=-1

\begin{table*}
  \centering
  \small
  \begin{tabular}{l  c c c c c c }
  \toprule
  & \textit{de} &
  \textit{it} &
  \textit{pl} &
  \textit{pt} &
  \textit{ru} &
  \textit{es}\\
  \midrule
  $\MI(G_a^\ell, A_a^\ell)$ & \textbf{0.0928} & \textbf{0.1316} & \textbf{0.0621} & \textbf{0.0933} & \textbf{0.0845} & \textbf{0.1111}\\
  $\MI(G_a^\ell, D_a^\ell)$ & \textbf{0.0410} & \textbf{0.0543} & \textbf{0.0273} & \textbf{0.0320} & \textbf{0.0664} & \textbf{0.0091} \\
  $\MI(G_a^\ell, I_a^\ell)$ & \textbf{0.0737} & \textbf{0.0543} & \textbf{0.0439} & \textbf{0.0687} & \textbf{0.0600} & \textbf{0.0358}\\
  $\MI(G_a^\ell, S_a^\ell)$ & \textbf{0.0343} & \textbf{0.0543} & \textbf{0.0258} & \textbf{0.0252} & \textbf{0.0303} & \textbf{0.0192} \\
  $\MI(G_a^\ell, C_a^\ell)$ & $<$ 0.001 &  N/A & $<$ 0.001  & N/A & $<$ 0.001  & N/A \\
  $\MI(G_a^\ell, N_a^\ell)$ & $<$ 0.001  &$<$ 0.001  & $<$ 0.001  & $<$ 0.001  & $<$ 0.001 & $<$ 0.001\\
\bottomrule
  \end{tabular}
    \caption{The mutual information (MI) between the grammatical genders of animate nouns and a) the adjectives used to describe those nouns (top row), b) the verbs that take those nouns as direct objects, as indirect objects, and as subjects (rows 2--4, respectively), and c) the cases and numbers of those nouns (rows 5 and 6, respectively) for \XX different gendered languages. Statistical significance (i.e., a $p$-value less than 0.05) is indicated using bold. MI values are not comparable across pairs of random variables.\looseness=-1}
  \label{tab:mi_significance_animate}
  \end{table*}

In \cref{tab:mi_significance_animate}, we provide $\MI(G_a^\ell,
A_a^\ell)$, $\MI(G_a^\ell, D_a^\ell)$, $\MI(G_a^\ell, I_a^\ell)$,
$\MI(G_a^\ell, S_a^\ell)$, $\MI(G_a^\ell, C_a^\ell)$, and
$\MI(G_a^\ell, N_a^\ell)$ for each language $\ell \in \{\textit{de},
\textit{it}, \textit{pl}, \textit{pt}, \textit{ru}, \textit{es}\}$. As
with inanimate nouns, we find that there is a statistically
significant relationship between the grammatical genders of animate
nouns and the adjectives used to describe those nouns. We also find
that there are statistically significant relationships between the
grammatical genders of animate nouns and the verbs that take those
nouns as direct objects, as indirect objects, and as subjects. Again,
the relationship for the
verbs that take those nouns as indirect objects involves a small
number of noun--verb pairs. As expected, we do not find any
statistically significant relationships for either case or
number.\looseness=-1

\begin{figure*}
  \centering
  \small
  \includegraphics[width=0.7\textwidth]{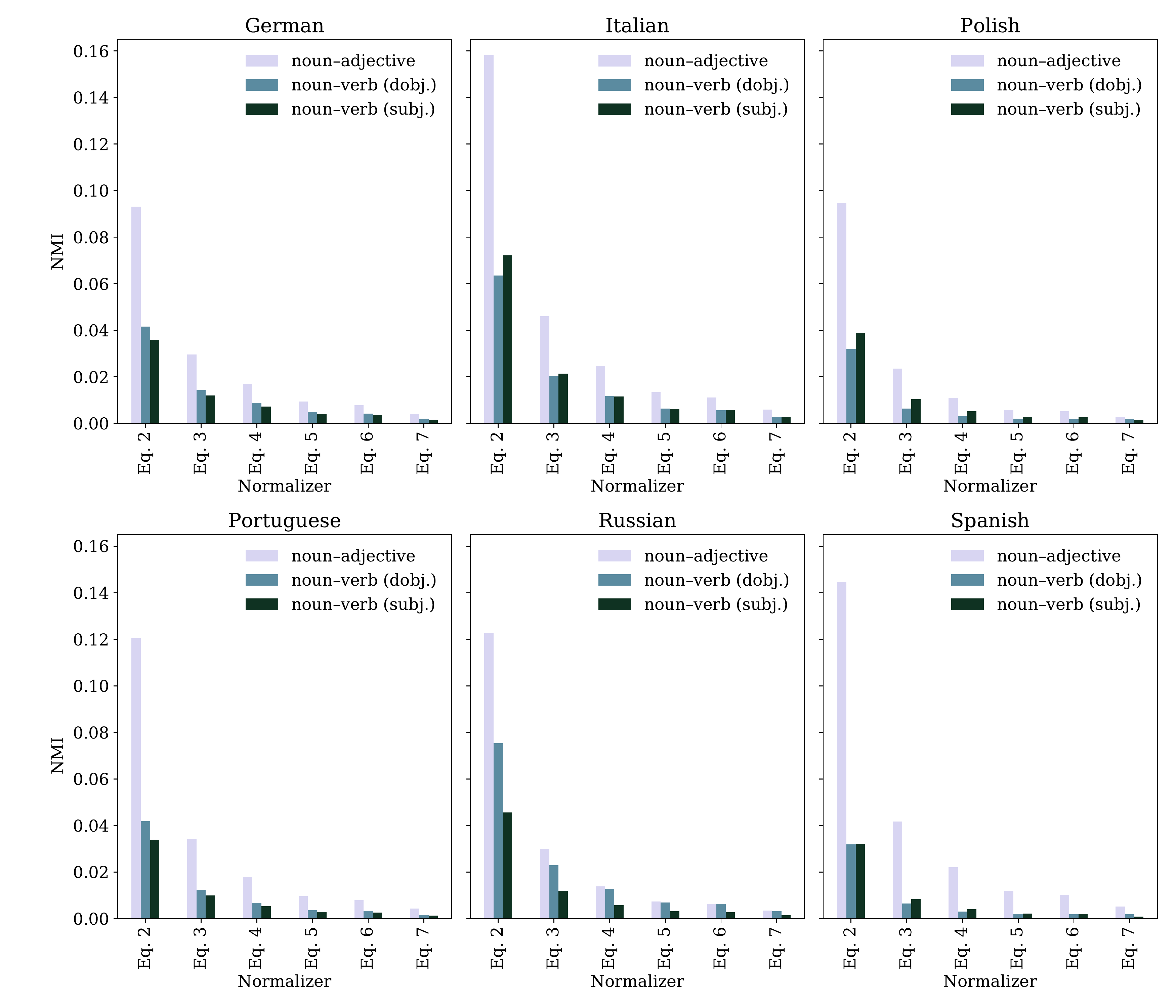}
  \caption{The normalized mutual information (NMI) between the
    grammatical genders of animate nouns and a) the adjectives used to
    describe those nouns b) the verbs that take those nouns as direct
    objects and as subjects for \XX different gendered languages. Each
    subplot contains six variants of~$\NMI(G_a^\ell, A_a^\ell)$,
    $\NMI(G_a^\ell,D_a^\ell)$, and $\NMI(G_a^\ell,S_a^\ell)$---one
    per normalizer---for a single language $\ell \in \{\textit{de},
    \textit{it}, \textit{pl}, \textit{pt}, \textit{ru},
    \textit{es}\}$.}
  \label{fig:nmi_animates}
\end{figure*}

\Cref{fig:nmi_animates} is analogous to \Cref{fig:nmi_inanimates}, in
that each subplot contains six variants of $\NMI(G_a^\ell, A_a^\ell)$,
$\NMI(G_a^\ell,D_a^\ell)$, and $\NMI(G_a^\ell,S_a^\ell)$, calculated
using normalizers that are analogous to those in \Cref{eq:norm1}
through \Cref{eq:norm6}, for a single language $\ell \in
\{\textit{de}, \textit{it}, \textit{pl}, \textit{pt}, \textit{ru},
\textit{es}\}$. (As with inanimate nouns, we omit
$\NMI(G_a^\ell,I_a^\ell)$ from each plot because of the small number
of noun--verb pairs involved.)
For $\ell \in \{\textit{de}, \textit{it}, \textit{pl},
\textit{pt}, \textit{es}\}$, $\NMI(G_i^\ell, A_i^\ell)$ is larger than
$\NMI(G_i^\ell,D_i^\ell)$ and $\NMI(G_i^\ell,S_i^\ell)$, regardless of
the normalizer. For $\ell \in \{\textit{it}, \textit{pl}\}$,
$\NMI(G_i^\ell,S_i^\ell)$ is larger than $\NMI(G_i^\ell,D_i^\ell)$;
for $\ell \in \{\textit{de}, \textit{pt}\}$, $\NMI(G_i^\ell,D_i^\ell)$
is larger than $\NMI(G_i^\ell,S_i^\ell)$; and
$\NMI(G_i^\textit{es},D_i^\textit{es})$ and
$\NMI(G_i^\textit{es},S_i^\textit{es})$ are roughly
comparable---again, all regardless of the normalizer. Meanwhile,
$\NMI(G_i^\textit{ru}, A_i^\textit{ru})$ is larger than
$\NMI(G_i^\textit{ru},D_i^\textit{ru})$ which is larger than
$\NMI(G_i^\textit{ru},S_i^\textit{ru})$ for the normalizers in
\Cref{eq:norm1} and \Cref{eq:norm2}, while $\NMI(G_i^\textit{ru},
A_i^\textit{ru})$ and $\NMI(G_i^\textit{ru},D_i^\textit{ru})$ are
roughly comparable and larger than
$\NMI(G_i^\textit{ru},S_i^\textit{ru})$ for the other five normalizers.\looseness=-1

\begin{figure*}
  \centering
  \small
  \includegraphics[width=0.7\textwidth]{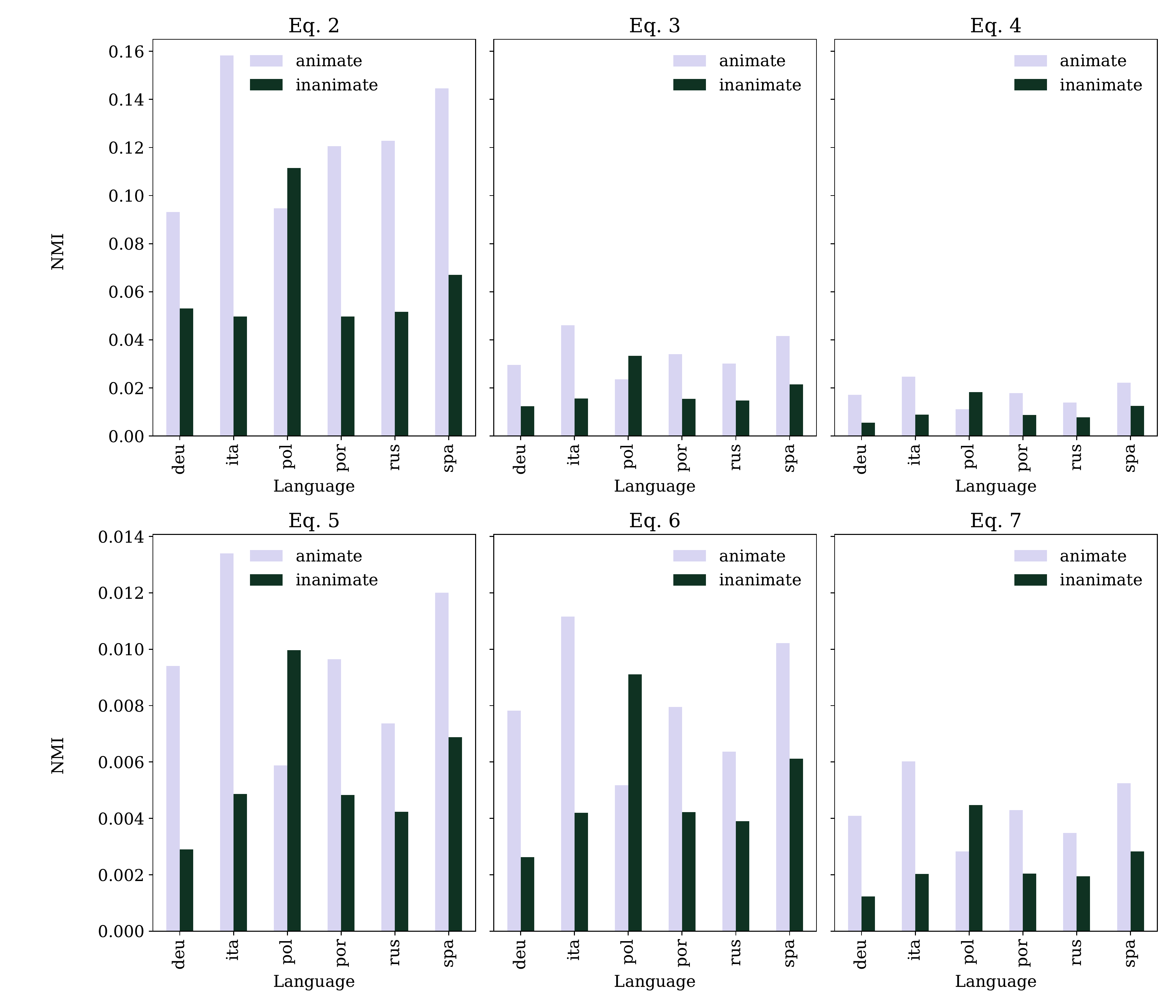}
   \caption{The normalized mutual information (NMI) between the
     grammatical genders of a) inanimate and b) animate nouns and the
     adjectives used to describe those nouns. Each subplot contains
     $\NMI(G^\ell_i, A_i^\ell)$ and $\NMI(G^\ell_a, A_a^\ell)$,
     calculated using a single normalizer, for each language $\ell \in
     \{\textit{de}, \textit{it}, \textit{pl}, \textit{pt},
     \textit{ru}, \textit{es}\}$.}
  \label{fig:nmi_both_adj}
\end{figure*}

\begin{figure*}
  \centering
  \small
  \includegraphics[width=0.7\textwidth]{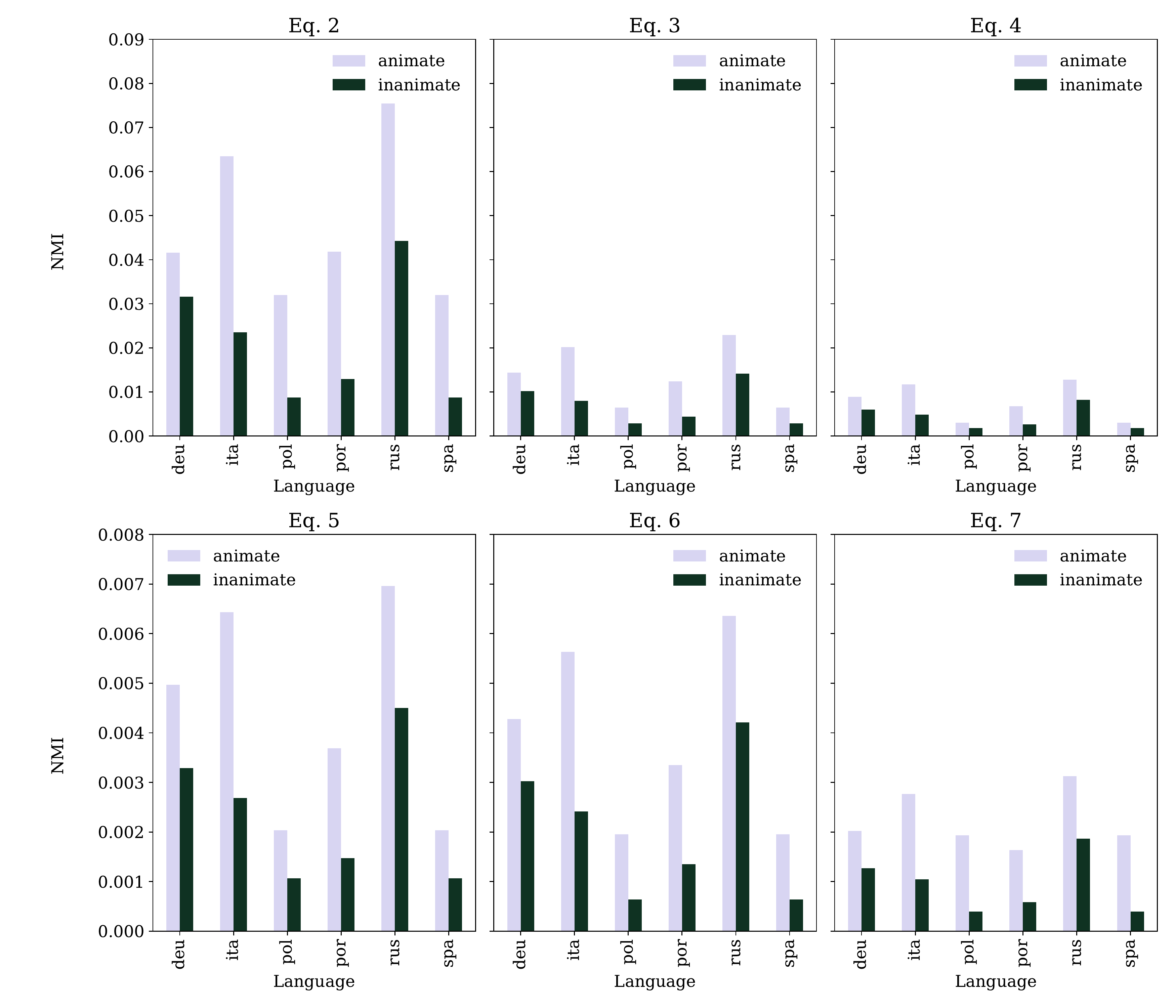}
   \caption{The normalized mutual information (NMI) between the
     grammatical genders of a) inanimate and b) animate nouns and the
     verbs that take those nouns as direct objects. Each subplot
     contains $\NMI(G^\ell_i, D_i^\ell)$ and $\NMI(G^\ell_a,
     D_a^\ell)$, calculated using a single normalizer, for each
     language $\ell \in \{\textit{de}, \textit{it}, \textit{pl},
     \textit{pt}, \textit{ru}, \textit{es}\}$.}
  \label{fig:nmi_both_dobj}
\end{figure*}

\begin{figure*}
  \centering
  \small
  \includegraphics[width=0.7\textwidth]{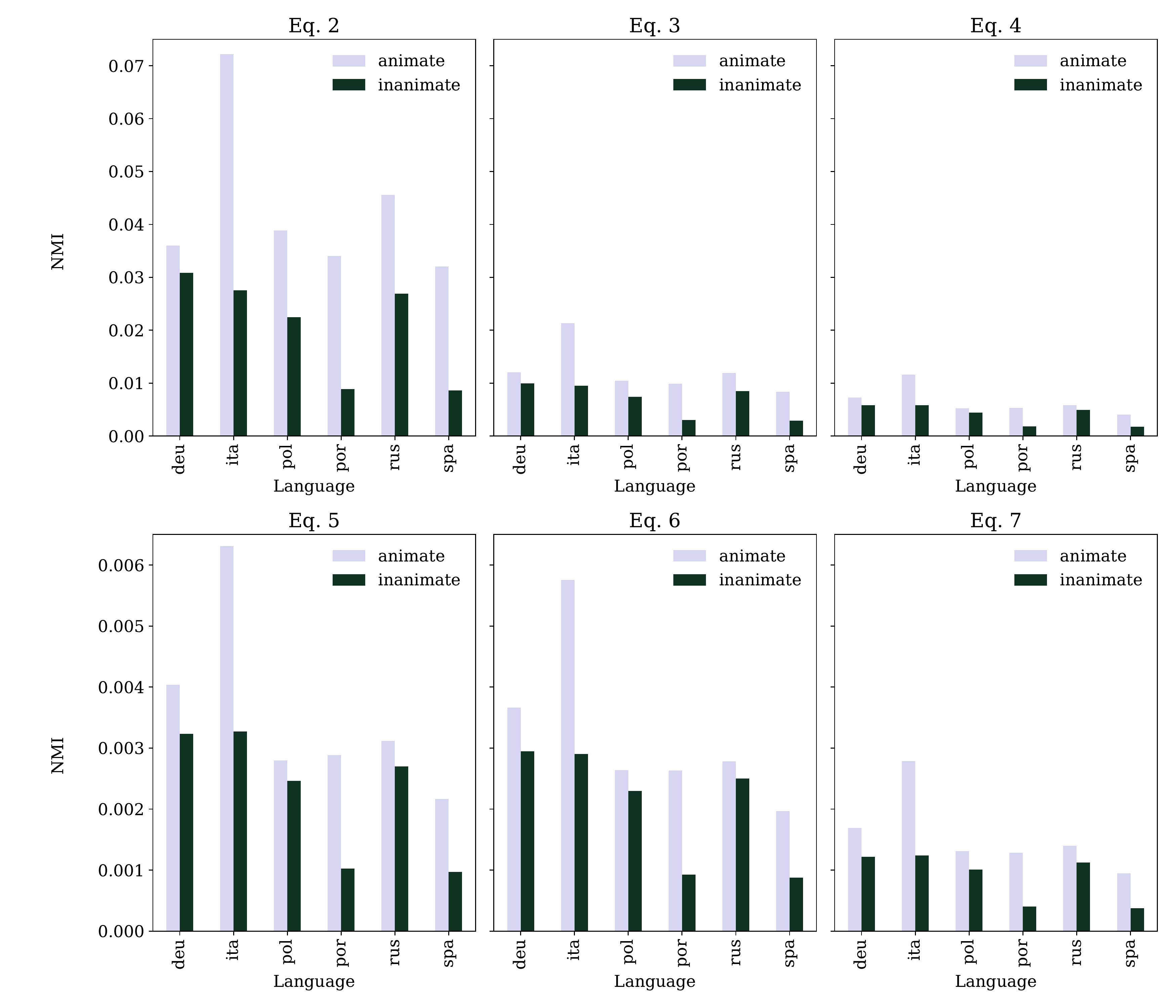}
   \caption{The normalized mutual information (NMI) between the
     grammatical genders of a) inanimate and b) animate nouns and the
     verbs that take those nouns as subjects. Each subplot contains
     $\NMI(G^\ell_i, S_i^\ell)$ and $\NMI(G^\ell_a, S_a^\ell)$,
     calculated using a single normalizer, for each language $\ell \in
     \{\textit{de}, \textit{it}, \textit{pl}, \textit{pt},
     \textit{ru}, \textit{es}\}$.}
  \label{fig:nmi_both_subj}
  \end{figure*}

Finally, each subplot in \Cref{fig:nmi_both_adj} contains
$\NMI(G_i^\ell, A_i^\ell)$ and $\NMI(G_a^\ell, A_a^\ell)$, calculated
using a single normalizer, for each for each language $\ell \in
\{\textit{de}, \textit{it}, \textit{pl}, \textit{pt}, \textit{ru},
\textit{es}\}$. Each subplot in \Cref{fig:nmi_both_dobj} analogously
contains $\NMI(G_i^\ell, D_i^\ell)$ and $\NMI(G_a^\ell, D_a^\ell)$,
while each subplot in \Cref{fig:nmi_both_subj} contains
$\NMI(G_i^\ell, S_i^\ell)$ and $\NMI(G_a^\ell, S_a^\ell)$. The NMI
values for animate nouns are generally larger than the NMI values for
inanimate nouns. The only exception is Polish, where
$\NMI(G_i^\textit{pl}, A_i^\textit{pl})$ is larger than
$\NMI(G_a^\textit{pl}, A_a^\textit{pl})$, regardless of the
normalizer.

\section{Discussion}

We find evidence for the presence of a statistically significant
relationship between the grammatical genders of inanimate nouns and
the adjectives used to describe those nouns for six different gendered
languages (specifically, German, Italian, Polish, Portuguese, Russian,
and Spanish). We also find evidence for the presence of statistically
significant relationships between the grammatical genders of inanimate
nouns and the verbs that take those nouns as direct objects, as
indirect objects, and as subjects. However, we caution against reading
too much into the relationship for the verbs that take those nouns as
indirect objects because of the small number of noun--verb pairs
involved. The effect sizes (operationalized as NMI values) for all of
these relationships are smaller than the effect sizes for animate
nouns. As expected, we do not find any statistically significant
relationships for either case or number.\looseness=-1

We emphasize that our findings complement, rather than supersede,
laboratory experiments, such as that of \citet{boroditsky2003sex}. We
use large-scale corpora and tools from NLP and information theory to
test for the presence of even relatively weak relationships across
multiple different gendered languages---and, indeed, the relationships
that we find have effect sizes (operationalized as NMI values) that
are small. In contrast, laboratory experiments typically focus on much
stronger relationships by tightly controlling experimental conditions
and measuring speakers' immediate, real-time speech
production. Moreover, although we find statistically
significant relationships, we do not investigate the characteristics
of these relationships. This means that we do not know whether they
are characterized by gender stereotypes, as argued by some cognitive
scientists, including \citet{boroditsky2003sex}. We also do not know
whether the relationships that we find are causal in nature. Because
MI is symmetric, our findings say nothing about whether the
grammatical genders of inanimate nouns \emph{cause} writers to choose
particular adjectives or verbs. We defer a deeper investigation of
these both of these avenues for future work.

Finally, we note that each of our tests can be viewed as a comparison
of the similarity of two clusterings of a set of items---specifically,
a ``clustering'' of nouns into grammatical genders and a
``clustering'' of the same nouns into, e.g., adjective
lemmata. Although (normalized) MI is a standard measure for comparing
clusterings, it is not without limitations (see, e.g.,
\citet{newman20} for an overview). For future work, we therefore
recommend replicating our tests using other information-theoretic
measures for comparing clusterings.\looseness=-1

\section*{Acknowledgments}

We thank Lera Boroditsky, Hagen Blix, Eleanor Chodroff, Andrei
Cimpian, Zach Davis, Jason Eisner, Richard Futrell, Todd Gureckis,
Katharina Kann, Peter Klecha, Zhiwei Li, Ethan Ludwin-Peery, Alec
Marantz, Arya McCarthy, John McWhorter, Sabrina J. Mielke, Elizabeth
Salesky, Arturs Semenuks, and Colin Wilson for discussions at various
points related to the ideas in this paper.

\appendix
\section{Appendix A: Counts}

Counts of the noun--adjective and noun--verb pairs for all six
gendered languages are in \cref{tab:inanimate_counts} (for inanimate
nouns) and \cref{tab:animate_counts} (for animate nouns).\looseness=-1

\begin{table*}
\centering
\small
\begin{tabular}{r cccccc}
\toprule
 & \textit{de} & \textit{it} & \textit{pl} & \textit{pt} & \textit{ru} & \textit{es} \\
 \midrule
\# noun--adj. tokens & 6443907 & 6246856 & 11631913 & 640558 & 32900200 & 3605439\\
\# noun--adj. types  & 770952 & 666656 & 640107 & 638774 & 1633963 & 368795 \\
\# noun types        & 10712 & 6410 & 5533 & 5672 & 9327 & 6157 \\
\# adj. types        & 4129 & 3607 & 4080 & 3431 & 11028 & 1907\\
\midrule
\# noun--verb (subj.) tokens & 3191030 & 1432354 & 2179396 & 1871941 & 6007063 & 1534211\\
\# noun--verb (subj.) types & 445536 & 292949 & 297996 & 337262 & 864480 & 376888\\
\# noun (subj.) types & 10741 & 6318 & 5522 & 5780 & 9129 & 7470 \\
\# verb types & 707 & 702 & 874 & 758 & 1803 & 875 \\
\midrule
\# noun--verb (dobj.) tokens & 3440922 & 2855037 & 3964828 & 4850012 & 6738606 & 2859135 \\
\# noun--verb (dobj.) types & 427441 & 393246 & 236849 & 541347 & 713703 & 576835\\
\# noun (dobj.) types & 10504 & 6407 & 4359 & 5896 &  8998 & 11567\\
\# verb types & 805 & 806 & 708 & 738 & 1539 & 9746 \\
\midrule
\# noun--verb (iobj.) tokens & 163935 & 71 & 54138 & 95009 & 1570273 & 56038\\
\# noun--verb (iobj.) types & 50133 & 53 & 18214 & 39738 & 300703 & 24830\\
\# noun (iobj.) types & 5520 & 59 & 2258 & 3757 & 8150 & 3574\\
\# verb types & 386 & 68 & 417 & 357 & 1816 & 464 \\
\midrule
\# noun--case tokens & 14681293 & N/A & 15300621 & N/A & 51641929 & N/A \\
\# noun--case types & 2252632 & N/A & 1465314 & N/A &5028075 & N/A \\
\# noun types & 11989 & N/A & 5839 & N/A & 9692& N/A \\
\# case types & 4 & 0 & 7 & 0 & 6 & 0 \\
\midrule
\# noun--number tokens &14681293 & 11588448& 15300621& 14631732& 51641929 & 5672790\\
\# noun--number types & 2252632& 1748927 & 1465314 & 2042626& 5028075 & 1034307\\
\# noun types &11989 & 7014 & 5839 & 6256 & 9692& 1593\\
\# number types & 2 & 2 & 2 & 2 & 2 & 2 \\
\bottomrule
\end{tabular}
\caption{Counts of the inanimate noun--adjective and noun--verb pairs for all \XX gendered languages.}
\label{tab:inanimate_counts}
\end{table*}

\begin{table*}
\centering
\small
\begin{tabular}{r cccccc}
\toprule
 & \textit{de} & \textit{it} & \textit{pl} & \textit{pt} & \textit{ru} & \textit{es} \\
 \midrule
\# noun--adj. tokens & 662760 & 818300 & 1137209 & 712101 & 3225932 & 387025\\
\# noun--adj. types & 99332 & 92424 & 97847 & 90865 & 264117 & 50173\\
\# noun types & 1998 & 1078 & 954 & 1006 & 2098 & 1320 \\
\# adj. types & 3587 & 3507 & 3836 & 3176 & 9833 & 1828\\
\midrule
\# noun--verb (subj.) tokens & 637801 & 399747 & 526894 & 456349 & 1516740 & 310569\\
\# noun--verb (subj.) types & 113308 & 77551 & 89819 & 89959 & 253150& 93586\\
\# noun (subj.) types & 2056 & 1066 & 969 & 1013 & 2020 & 1477\\
\# verb types & 707 & 702 & 874 & 758 & 1799 & 874 \\
\midrule
\# noun--verb (dobj.) tokens & 321400 & 388187 & 456824 & 527259 & 494534 & 850234\\
\# noun--verb (dobj.) types & 60760 & 55574 & 76348 & 92220 & 118818 & 85235 \\
\# noun (dobj.) types & 1901 & 1025 & 867 & 1028 & 1912 & 1023\\
\# verb types & 804 & 805 & 724 & 737 & 1535 & 745 \\
\midrule
\# noun--verb (iobj.) tokens & 51359 & 7 & 43187 & 23139 & 518540 & 23955\\
\# noun--verb (iobj.) types & 17804 & 6 & 8440 & 110185 & 11353 &  9586  \\
\# noun (iobj.) types & 1149 & 6 & 628 & 773 & 1858 & 947 \\
\# verb types & 378 & 6 & 411 & 340 & 1769 & 456 \\
\midrule
\# noun--case tokens & 1926614 & N/A & 1907688 & N/A & 6357089 & N/A \\
\# noun--case types & 390672   & N/A & 299511 & N/A & 987420 & N/A  \\
\# noun types       & 2292     & N/A & 1024 & N/A &2194 &  N/A \\
\# case types &            4 & 0 & 7 & 0 &6 & 0  \\
\midrule
\# noun--number tokens & 1926614 & 1801285 &1907688 &  1931315 & 6357089 & 786177 \\
\# noun--number types & 390672 & 306968 & 299511 & 356352 & 987420& 200785\\
\# noun types &  2292 & 1135 &1024  & 1072 & 2194  &1593 \\
\# number types & 2 & 2 & 2 & 2 &2 & 2 \\
\bottomrule
\end{tabular}
\caption{Counts of the animate noun--adjective and noun--verb pairs for all \XX gendered languages.}
\label{tab:animate_counts}
\end{table*}

\comment{
\section{Appendix B: Plots}

\begin{figure*}
  \centering
  \small
  \includegraphics[width=0.7\textwidth]{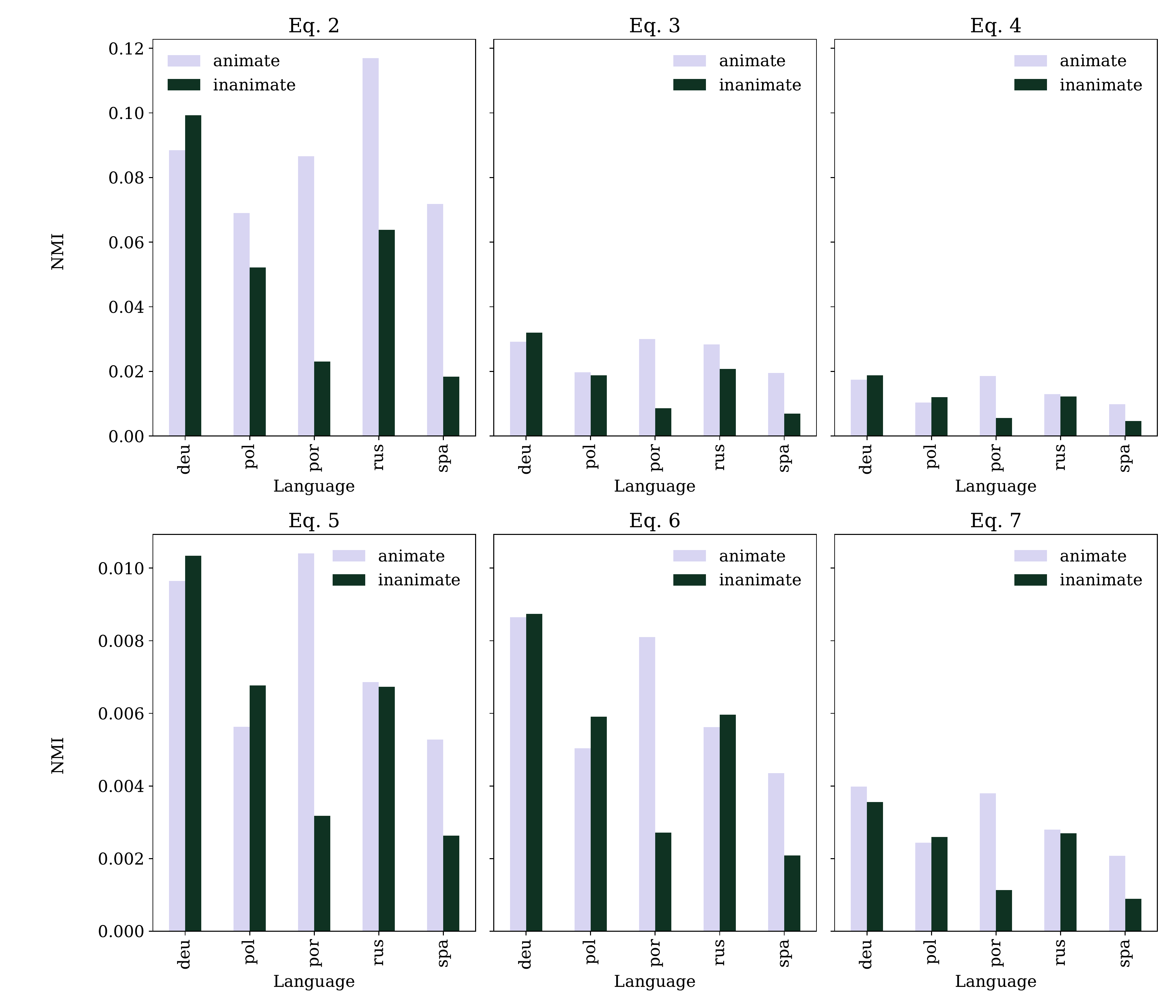}
   \caption{The normalized mutual information (NMI) between the
     grammatical genders of a) inanimate and b) animate nouns and the
     verbs that take those nouns as indirect objects. Each subplot
     contains $\NMI(G^\ell_i, D_i^\ell)$ and $\NMI(G^\ell_a,
     D_a^\ell)$, calculated using a single normalizer, for each
     language $\ell \in \{\textit{de}, \textit{pl}, \textit{pt},
     \textit{ru}, \textit{es}\}$; we omit Italian because we do not
     find a statistically significant relationship.\looseness=-1}
  \label{fig:nmi_both_iobj}
\end{figure*}
}

\clearpage

\bibliography{gender-mi}
\bibliographystyle{acl_natbib}

\end{document}